\documentclass{article}
\usepackage{graphicx}
\usepackage{amssymb}
\usepackage{amsmath}
\usepackage{epstopdf}
\usepackage[tight,footnotesize]{subfigure}
\usepackage{stfloats}
\usepackage{microtype}
\usepackage{units}
\usepackage{algorithm}
\usepackage[noend]{algpseudocode}
\usepackage{url}
\usepackage{bm}
\usepackage{multirow}

\usepackage{float}
\usepackage{adjustbox}
\usepackage{multirow}

\usepackage{booktabs} 
\usepackage{verbatim} 

\begin{document}
\title{Can automated smoothing significantly improve benchmark time series classification algorithms?}

\author{James Large \and Paul Southam \and Anthony Bagnall}
\date{November 2018}
%
%

\maketitle              

\begin{abstract}
tl;dr: no, it cannot, at least not on average on the standard archive problems. We assess whether using six smoothing algorithms (moving average, exponential smoothing, Gaussian filter, Savitzky-Golay filter, Fourier approximation and a recursive median sieve) could be automatically applied to time series classification problems as a preprocessing step to improve the performance of three benchmark classifiers (1-Nearest Neighbour with Euclidean and Dynamic Time Warping distances, and Rotation Forest). We found no significant improvement over unsmoothed data even when we set the smoothing parameter through cross validation. We are not claiming smoothing has no worth. It has an important role in exploratory analysis and helps with specific classification problems where domain knowledge can be exploited. What we observe is that the automatic application does not help and that we cannot explain the improvement of other time series classification algorithms over the baseline classifiers simply as a function of the absence of smoothing.

\end{abstract}

\section{Introduction}
Time Series Classification (TSC) is differentiated from standard classification by the fact that the ordering of the attributes may be important in finding discriminatory features. Standard vector classifiers such as rotation forest and standard time dependent approaches such dynamic time warping with 1-NN are strong benchmark algorithms to compare against the range of bespoke TSC algorithms that have been proposed in recent years. Some of these achieve impressive performance and are significantly better than the benchmarks. Nevertheless, there has always been a suspicion that sensible standard preprocessing of the data would perhaps increase the accuracy of benchmark classifiers and that would make at least some of the bespoke algorithms redundant~\cite{hu13realistic}. Broadly speaking, there are four types of preprocessing that may improve classifier performance: normalisation; smoothing; dimensionality reduction; and discretization. We address the question of whether smoothing series can significantly improve the accuracy of benchmark classifiers. Smoothing is the process of reducing the noise in the series to make patterns in the data more apparent and is generally used as part of an exploratory analysis.

It is important to stress we are only concerned with class independent noise, since class dependent noise is possibly useful as a discriminatory feature. This is where we diverge from the majority of signal processing research into noise modeling and reduction. We are not necessarily trying to ``clean up" a signal. Instead, we are trying to remove artifacts that may confound the classifier.

We test whether six smoothing algorithms improve three base classifiers. These are described in detail in Section~\ref{sec:background}. It is clearly important to set the parameters of the algorithm when smoothing so that it is relevant to a specific problem. Because we are attempting to smooth to improve classification, we set parameters through cross validation on the train data using the base classifier we are testing. The experimental design is described in Section~\ref{sec:results}. Our experiments address the following two questions.
\begin{enumerate}
\item Does smoothing with default parameters increase the accuracy of benchmark classifiers?
\item Can we learn smoothing parameters on the train data to significantly improve  benchmark TSC algorithms?
\end{enumerate}

A priori, we believed it unlikely that systematic smoothing would improve accuracy over the diverse data sets in the archive, since many of the series have very little noise. However, we thought that supervised smoothing, where no smoothing was an option, would improve performance albeit at the large computational cost of the parameter search. Our results, presented in Section~\ref{sec:results} show that in fact smoothing makes very little difference, even when supervised. We discuss these results in Section~\ref{sec:conclusions}.

\section{Background}
\label{sec:background}
\subsection{Time Series Classification}

A large number of new classification problems have been proposed in the last ten years. While not exhaustive by itself, it is important to evaluate new algorithms against sensible benchmark classifiers on standard test problems in order to ascertain the usefulness of new research. The UEA-UCR archive is a widely used archive of test problems~\cite{UCRWeb}.  The archive is a continually growing collection of real valued TSC datasets\footnote{http://www.timeseriesclassification.com} which come from a range of different domains and have a range of characteristics, in terms of size, number of classes, imbalances, etc. Most TSC publications benchmark against a 1-NN classifier using either Euclidean distance (ED) or Dynamic Time Warping (DTW) distance.  DTW compensates for potential misalignments amongst series of the same class. DTW has a single parameter, the maximum warping window, and DTW performs significantly better when this parameter is set through cross validation. A recent comparative study~\cite{bagnall17bakeoff} found that the classifier rotation forest~\cite{rodriguez06rotf} was also a strong benchmark. It is able to discover relationships in time through the internal principle component transformation it uses, and is not significantly worse than DTW with window set through cross validation (DTWCV henceforth). The same study compared 22 TSC algorithms on 85 of the UEA-UCR archive data and found that just nine out of twenty two TSC algorithms were significantly more accurate than both a rotation forest and DTW classifier. Some of these TSC algorithms are highly complex and both memory and computationally expensive. A case was made that the superior algorithms achieved higher accuracy because the representation they use allows for the detection of discriminatory features that the benchmarks cannot find. This was further demonstrated on the archive and through data simulation~\cite{lines16hive}. We wish to test whether simple preprocessing can significantly improve the benchmarks and hence narrow the gap between DTW and rotation forest and the nine significantly better TSC algorithms.

\subsection{Time Series Smoothing}

Given a time series $T=<t_0,\ldots,t_{m-1}>$, a smoothing function produces a new series $S=<s_0,\ldots,s_{p-1}>$, where $p \leq m$ (we index from zero to make the equations simpler). Most algorithms employ a sliding window, of length $w$, along the series, resulting in a series of length $p=m-w$. The simplest form of smoothing is to take the {\bf moving average (MA)}.

$$s_j= \frac{\sum_{i=j-w}^{j} t_i}{w} \;\;\;\text{for} \;\;j=w \ldots m-1,$$
where $w$ is the single parameter, window size. {\bf Exponential smoothing (EXP)} is a generalisaton of moving average smoothing that assigns a decaying weight to each element rather than averaging over a window.

$$s_0= t_0 \;\;\;\text{and}\;\;\;
s_j= \alpha \cdot t_j+(1-\alpha)  \cdot t_{j-1}$$

where $0 \leq \alpha \leq 1$. For consistency with other smoothing algorithms, EXP is often given a window size $w$, then the decay weight is set as
$\alpha=\frac{2}{w+1}$.

A {\bf Gaussian filter (GF)} applies a fixed convolution over a window

$$s_j= \sum_{i=j-w}^{j} t_i \cdot c_{i}, $$
where the convolution values $c_{i}$ are derived from a standard normal distribution over the window $w$, the single parameter.

Like GF, the {\bf Savitzky-Golay (SG)} filtering method is a convolutional method of smoothing. Instead of using a fixed convolution, it estimates a different convolution on each window based on local least-squares polynomial approximation.

$$s_j= \sum_{i=j-w}^{j} t_i \cdot c_{i,j} $$
Since its initial introduction \cite{Savitzky1964}, it has been used successfully and pervasively across many signal processing domains for different purposes, particularly in chemometrics \cite{Betta2015,Fernandes2017}. SG has two parameters, window size $w$ and  polynomial order $n$. For accessible explanations of how the polynomial coefficients are calculated, we refer the reader to \cite{Schafer2011}.

{\bf Discrete Fourier Approximation (DFT)} smooths the series by first transforming into the frequency domain, discarding the high frequency terms, then transforming back to the time domain. DFT has a single parameter, $r$, the proportion of Fourier terms to retain.

The {\bf Recursive Median Sieve (SIV)} is a one-dimensional recursive median filter \cite{bangham1988} that filters the data by removing extrema of specific scales.

\begin{figure}[!h]
	\centering
		\includegraphics[width=.75\textwidth]{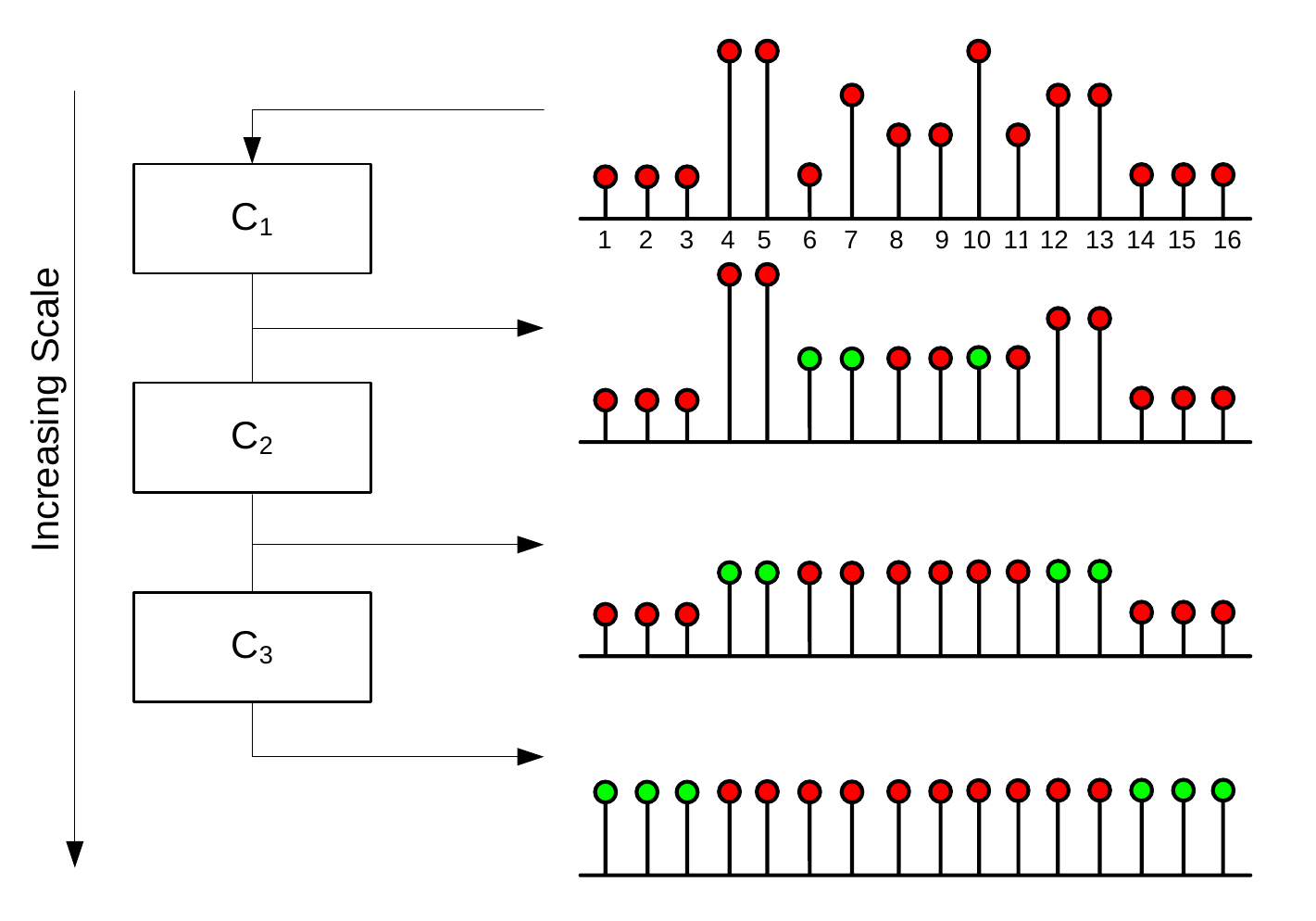}
	\caption{An example sieve decomposition of a 1D signal. Green vertices are the vertices affected at each scale level.}
	\label{fig:Msievegraph}
\end{figure}

The sieve performs a decomposition removing extrema (both maxima and minima) at different scales as shown in Figure \ref{fig:Msievegraph}. At the scale $c_1$ the maxima and minima at points 6,7 and 10 are smoothed to equal the nearest value of the neighbours. At scale $c_2$ the pairs at (4,5) and (12,13) are smoothed. At the highest scale, the series is uniform. The sieve takes in a single parameter, $c$, which is the scale to smooth the signal to.

\begin{table}[!h]
\caption{The parameter spaces searched for each filtering method over the course of our experiments (default value in bold). $m$ is the series length. For Savitzky-Golay (SG), all combinations of $w,n$ are searched where $w > 2n$. }
\centering
\label{tab:paraSpaces}

\begin{tabular}{l|l}
\hline
Method & Parameters and default values in bold \\
\hline
Moving Average (MA) &  $w \in \{2,3,{\bf 5},10,25,50,100,\sqrt{m},\log_2(m)\}$ \\
Exponential Smoothing (EXP) & $w \in \{2,3,{\bf 5},10,25,50,100,\sqrt{m},\log_2(m)\}$  \\
Gaussian Filtering (GF) &  $w \in \{2,3,{\bf 5},10,25,50,100,\sqrt{m},\log_2(m)\} $ \\
\multirow{ 2}{*}{Savitzky-Golay (SG)} &  $w \in \{{\bf 5},9,17,33,65 \}$ \\
&					 $n\in \{ {\bf 2},3,4,8,16,32 \}$ \\
Fourier Approximation (DFT) & $ r \in\{0.01,0.05,{\bf 0.1},0.25,0.5,\log_2(m) /m \}$ \\
Sieve (SIV) &   $c \in \{\frac{1}{15}\cdot \log_{10}(m), \ldots,{\bf \frac{5}{15}\cdot \log_{10}(m)}, \ldots,\log_{10}(m)\}$ \\[2pt] 

\hline
\end{tabular}
\end{table}

For each of pair of filter+classifier combination, we perform 10 stratified random resamples of each data set and report the average results across those resamples. The first resample, fold 0, is always the exact train/test split published on the UCR-UEA archive, to allow for easier comparison.  To avoid ambiguity, we stress that in all cases the training of a classifier, including any parameter tuning and model selection required, is performed independently on the train set of a given fold, and the trained classifier is evaluated exactly once on the corresponding test set. We conduct 10 resamples on 76 of the 85 UCR archive TSC problems. We have omitted the largest problems due to time constraints. We average test accuracy over the 10 resamples, then present results in critical difference diagrams, which display the average ranks of the classifiers over all problems and group classifiers into cliques, within which there is no significant difference. For each resample, we perform a 10 fold cross validation (CV) on that resamples' train data to find smoothing parameters. Our code\footnote{https://bitbucket.org/TonyBagnall/time-series-classification} reproduces the splits used in this evaluation exactly, and full, reproducible results are available\footnote{http://research.cmp.uea.ac.uk/SmoothingAALTD18/}. For all smoothing algorithms except the sieve, we used the standard MATLAB implementations and performed the smoothing and classification in separate stages. The default parameters given in Table~\ref{tab:paraSpaces} are those of the Matlab implementations. The sieve is implemented in C and was similarly isolated from the classification stage.

We use three baseline classifiers. 1-NN with Euclidean distance is a  weak baseline, but it is still frequently used in research. 1-NN with DTW is the most common benchmark, although it is important to set the window through cross validation~\cite{ratanamahatana05threemyths}. This is computationally expensive, although we use the DTW version described in~\cite{tan18efficientDTW} which speeds up the calculation by orders of magnitude. All the UEA-UCR data are normalised. For consistency, we renormalise each series after smoothing.

\section{Results}
\label{sec:results}
\begin{figure}[!htb]
\begin{center}
\begin{tabular}{cc}
       \includegraphics[width=.49\textwidth, trim={2cm 7cm 2cm 4cm},clip]{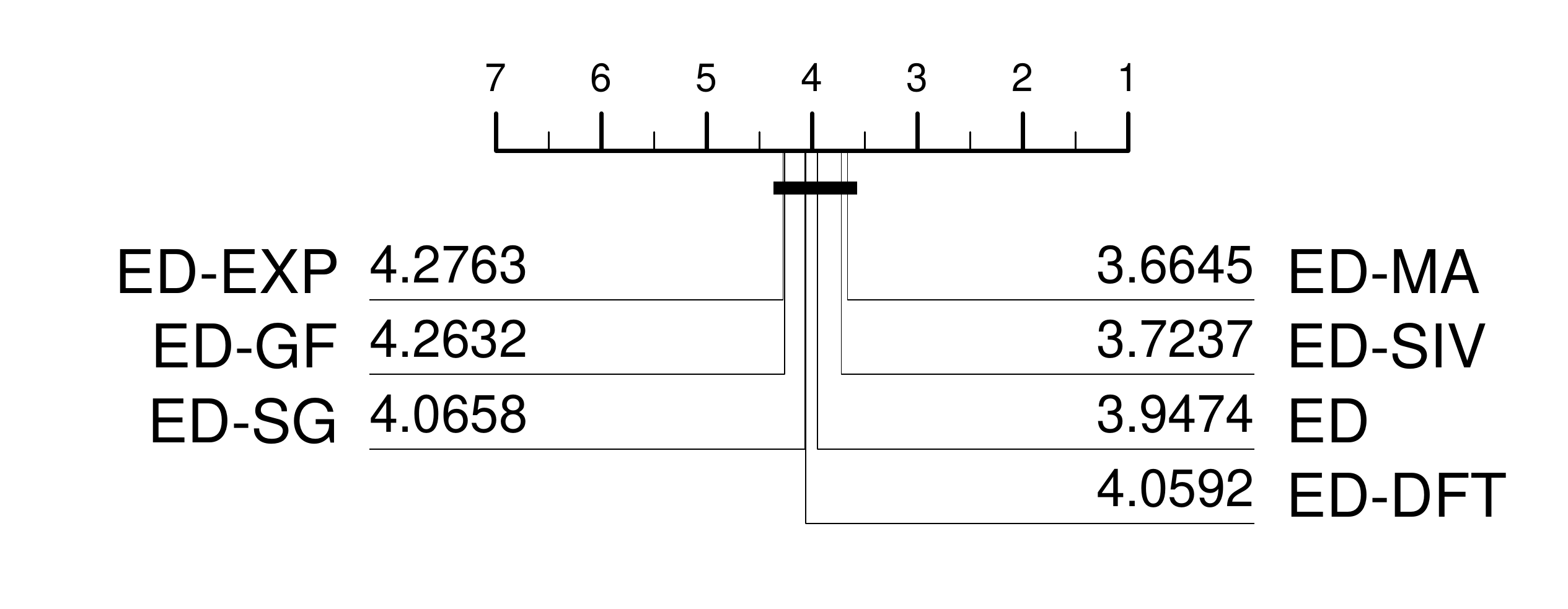} &
       \includegraphics[width=.49\textwidth, trim={2cm 7cm 2cm 4cm},clip]{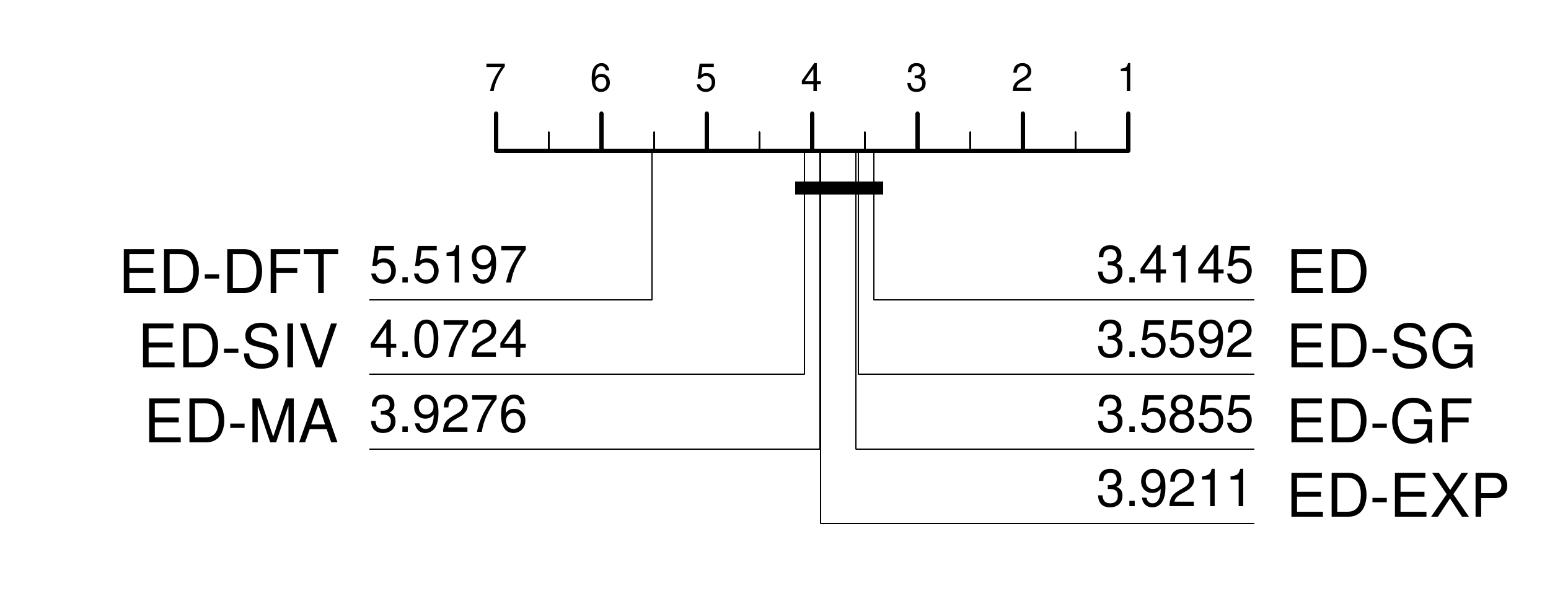}\\
(a) & (b)        \\
       \includegraphics[width=.49\textwidth, trim={0cm 6cm 0cm 4cm},clip]{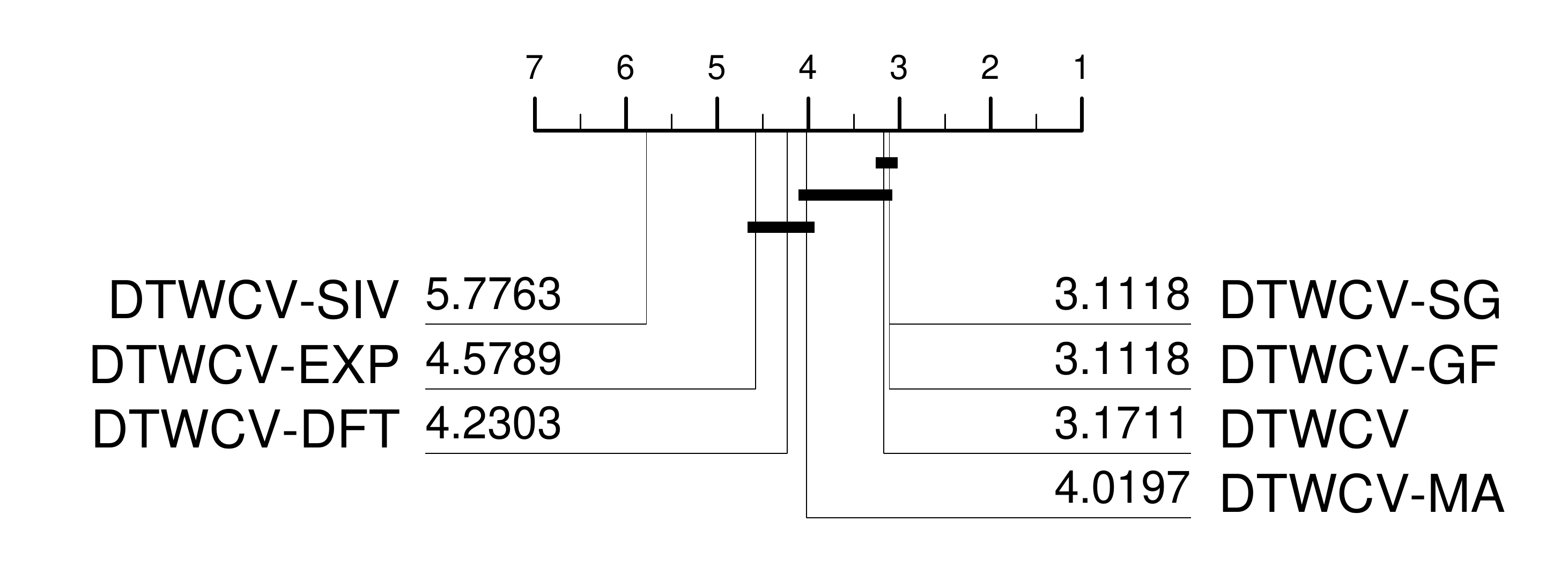} &
       \includegraphics[width=.49\textwidth, trim={0cm 6cm 0cm 4cm},clip]{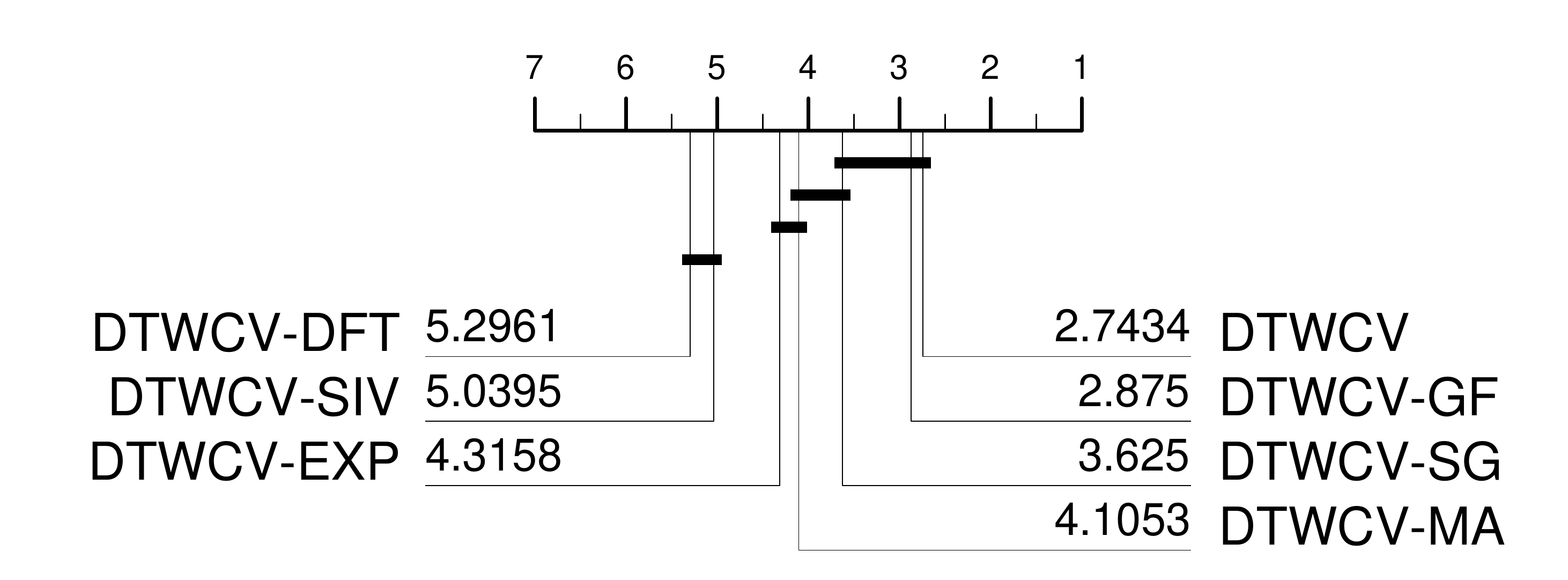}\\
(c) & (d)        \\
       \includegraphics[width=.49\textwidth, trim={2cm 7cm 2cm 4cm},clip]{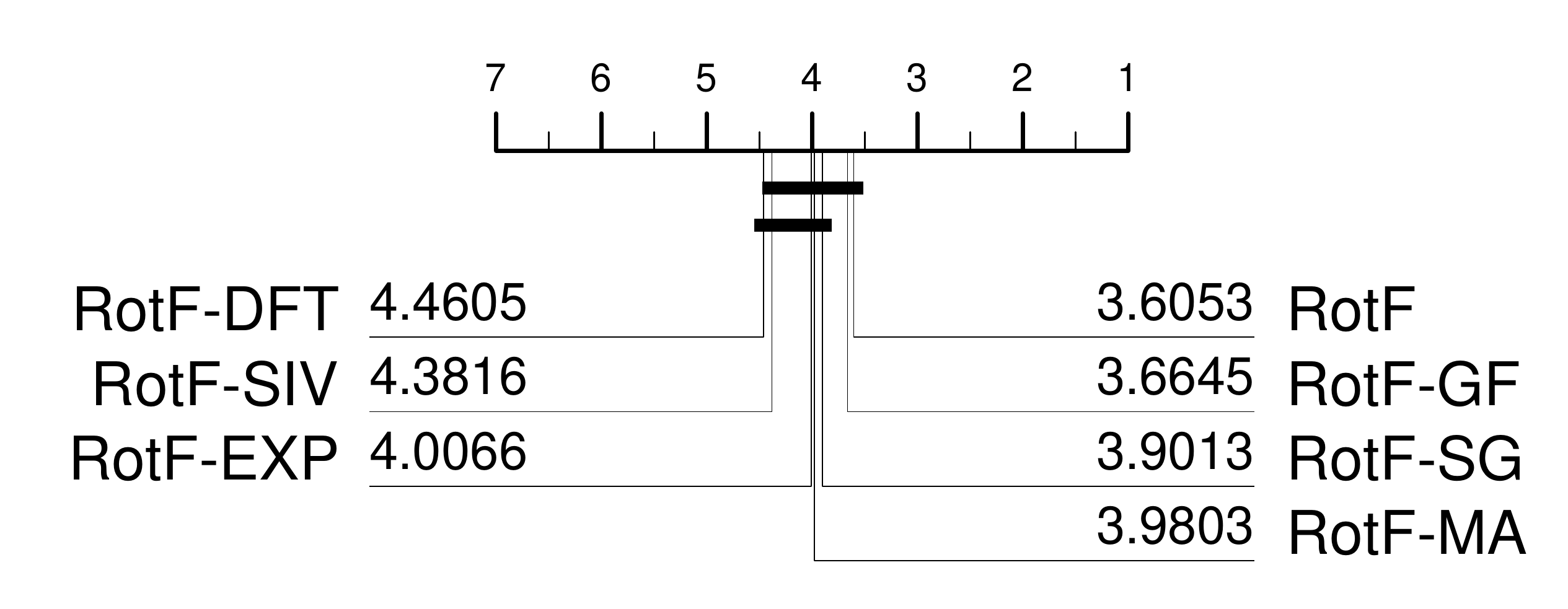} &
       \includegraphics[width=.49\textwidth, trim={2cm 7cm 2cm 4cm},clip]{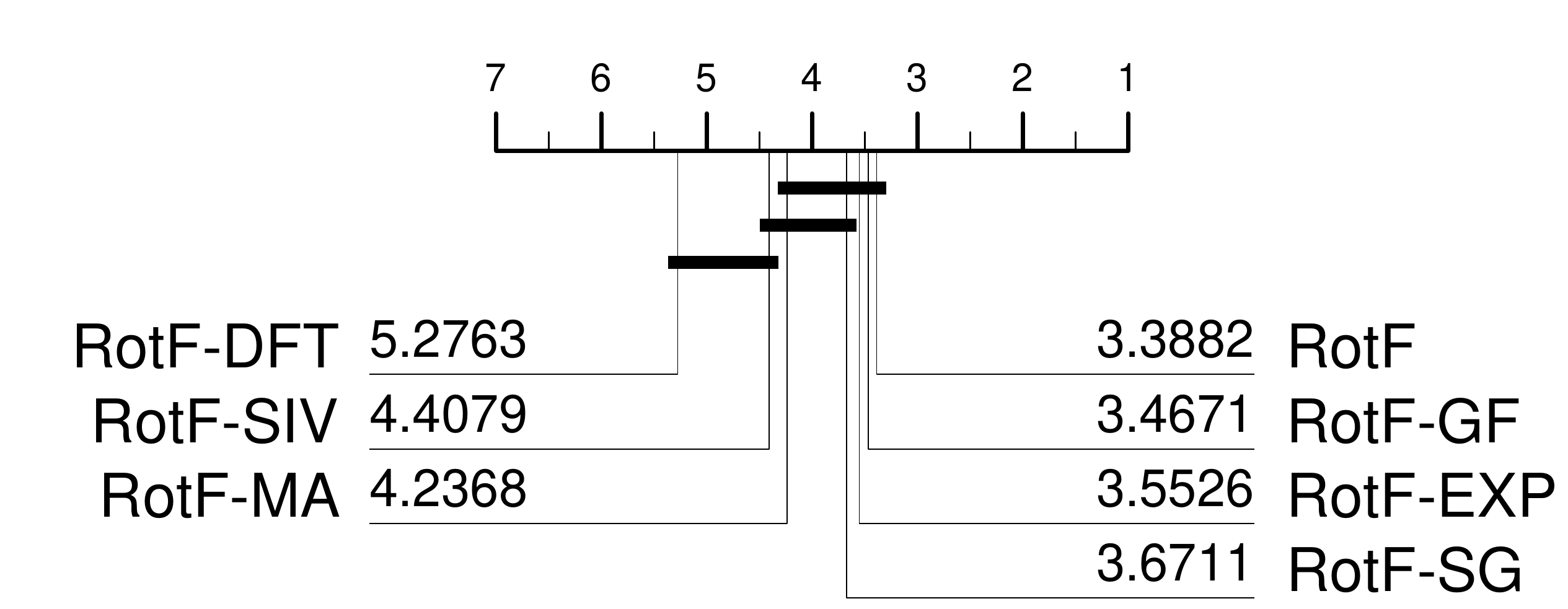}\\
(e) & (f)
\end{tabular}
\end{center}
\caption{Average ranks on 76 UEA-UCR problems without smoothing  (left) and with six types of smoothing described in Section~\ref{sec:background} (right), using the default parameters given in Table~\ref{tab:paraSpaces} and tuned over the parameter range given in Table~\ref{tab:paraSpaces} for Euclidean distance ((a) and (b)), dynamic time warping ((c) and (d)) and rotation forest ((e) and (f)).}
\label{fig:smooth}
\end{figure}

In Figure~\ref{fig:smooth} we present results for three baseline classifiers with both default smoothing and tuned smoothing.
For all three classifiers, smoothing of any kind provides no benefit. Tuning provides no benefit over using default values, and in many cases makes things worse due to overfitting.

For all six experiments, the classifiers built on unsmoothed data are in the top clique. For four of the experiments, the unsmoothed classifier is the highest ranked. Setting the parameter through cross validation is if anything worse than using a default parameter. Given the order of magnitude more computation required to tune these parameters, this is surprising, particularly as {\em no smoothing} was one of the options. Further analysis shows that {\em no smoothing} was selected approximately 25\% of the time. This could be an indication that the archive data are simply not suited to smoothing.

\section{Analysis}
\begin{figure}[!htb]
\begin{center}
       \includegraphics[width=0.9\textwidth, trim={2cm 10cm 2cm 10cm},clip]{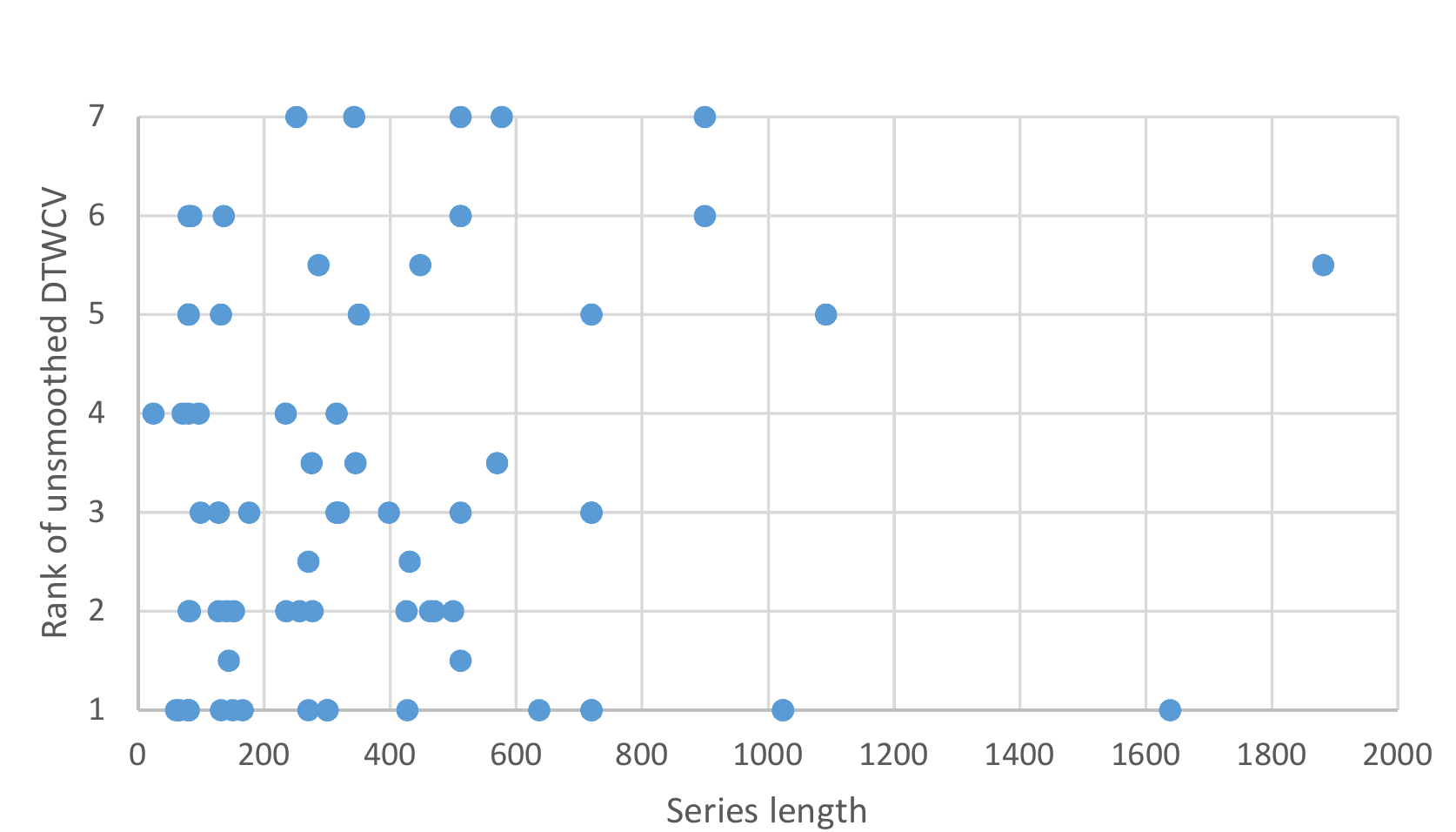}
\end{center}
\caption{Ranks on 76 UEA-UCR problems of DTWCV compared to six untuned smoothing versions plotted against series length. }
\label{fig:dtwSeriesLength}
\end{figure}

\begin{figure}[!htb]
\begin{center}
       \includegraphics[width=0.9\textwidth, trim={2cm 10cm 2cm 10cm},clip]{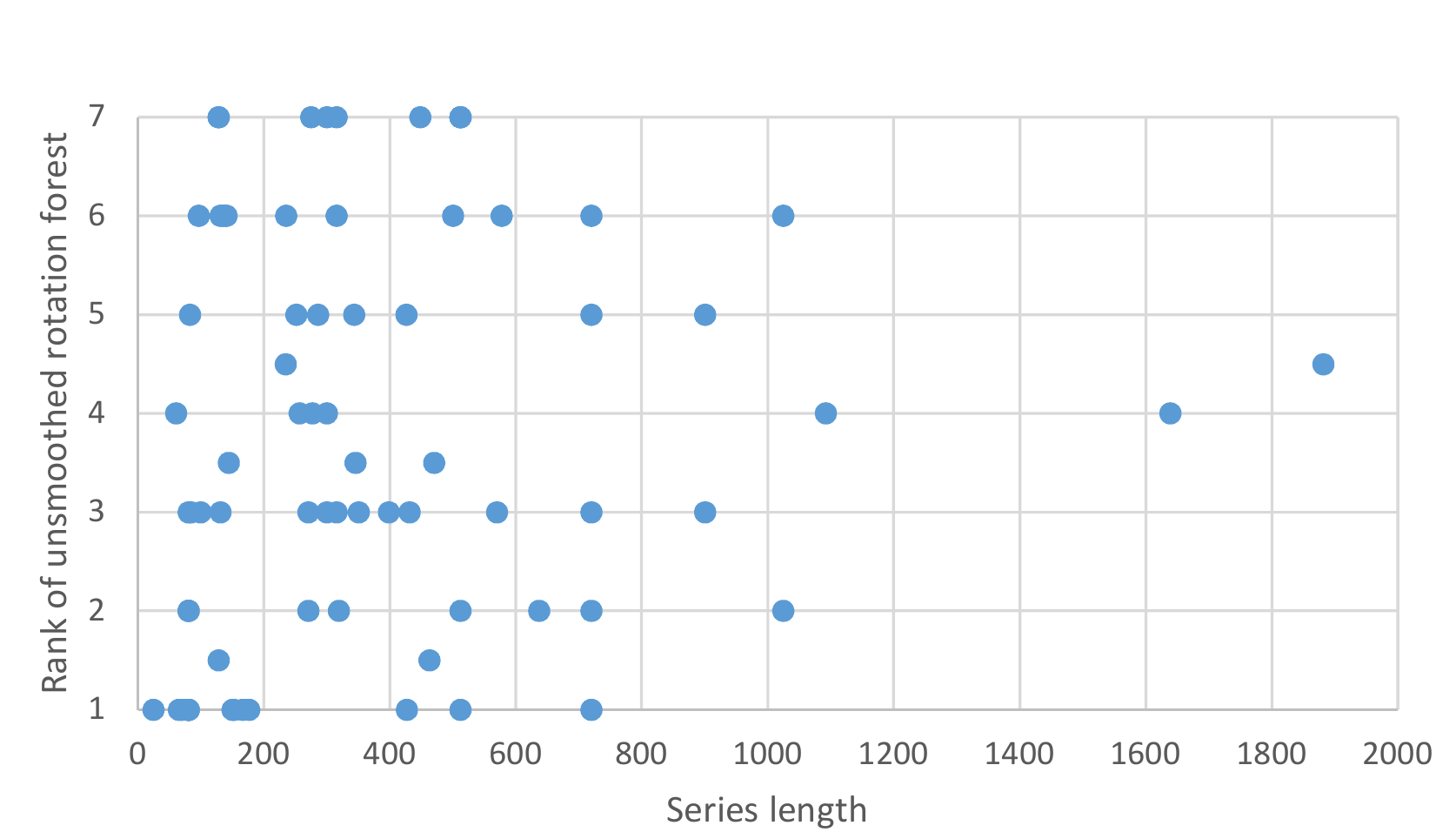}
\end{center}
\caption{Ranks on 76 UEA-UCR problems of rotation forest compared to six untuned smoothing versions plotted against series length. }
\label{fig:rotfSeriesLength}
\end{figure}

We examine whether there are any characteristics of the data that could help determine whether any of the six types of smoothing would improve performance. We would expect that smoothing might be more useful for longer series. Figures~\ref{fig:dtwSeriesLength} and~\ref{fig:rotfSeriesLength} show the scatter plot of length against classifier rank for DTWCV and rotation forest. There is no obvious relationship between the performance of the unsmoothed classifier and series length.


\section{Conclusion}
\label{sec:conclusions}
It has long been a suspicion of many researchers in this field that much of the improvement seen in complex TSC algorithms could equally be achieved with comparatively simple preprocessing. Our experiments indicate for the case of smoothing, this is not true. We have taken six very popular smoothing algorithms and applied them using sensible default parameters and using extensive extra computation to discover optimal parameters through cross validation. We have found no significant difference between smoothed and unsmoothed classification with three benchmarks. The nature of the UCR data may explain this to a degree: the data from problems such as image processing will have less noise than, for example, financial data. We are not claiming that smoothing has no role to play in TSC, merely that the application of smoothing does not on average improve the performance of baselines and that the absence of smoothing cannot explain the performance of algorithms that outperform the baselines. \\

\noindent {\bf Acknowledgement.} This work is supported by the UK Engineering and Physical Sciences Research Council (EPSRC)  [grant number EP/M015807/1] and
Biotechnology and Biological Sciences Research Council [grant number
BB/M011216/1]. The
experiments were carried out on the High Performance
Computing Cluster supported by the Research and
Specialist Computing Support service at the University
of East Anglia and using a Titan X Pascal donated by
the NVIDIA Corporation.

\end{document}